\documentclass[letterpaper,10pt,conference]{axisconf}
\IEEEoverridecommandlockouts
\usepackage[T1]{fontenc}

\usepackage{amsmath,amssymb,graphicx,url,array}
\newcommand{\clip}{\operatorname{clip}}
\title{\vspace{-12pt}\LARGE\bf What Matters in Quality-Aware VLA Pretraining:\\A Benchmark and Empirical Study}
\author{\parbox{\dimexpr\textwidth-4pt\relax}{\centering
{\fontsize{10.5}{13}\selectfont
Mingxuan Yan\textsuperscript{1},
Litian Gong\textsuperscript{3},
Dihong Huang\textsuperscript{2,4},
Zhixu Li\textsuperscript{3},
Tianyu Zhang\textsuperscript{3},
Wenqian Zhang\textsuperscript{3},\\
Hefeifei Jiang\textsuperscript{3},
Peihao Li\textsuperscript{5},
Jian Zhang\textsuperscript{6},
Mengfei Zhao\textsuperscript{2},
Yikai Tang\textsuperscript{4},
Hai Zhai\textsuperscript{2},\\
Zehao Wang\textsuperscript{1},
Ruijian Liang\textsuperscript{2},
Yanjia Huang\textsuperscript{6},
Lin Shao\textsuperscript{7},
Changliu Liu\textsuperscript{4},\\
Kaiyu Hang\textsuperscript{8},
Zhiwen Fan\textsuperscript{6},
Masayoshi Tomizuka\textsuperscript{5},
Jianfei Yang\textsuperscript{9},
and Jiachen Li\textsuperscript{1,\ensuremath{\ddagger}}\par}
\vspace{6pt}
{\fontsize{8.5}{11}\selectfont
\textsuperscript{1}Georgia Institute of Technology\quad
\textsuperscript{2}Axis Robotics\quad
\textsuperscript{3}University of California, Riverside\\
\textsuperscript{4}Carnegie Mellon University\quad
\textsuperscript{5}University of California, Berkeley\quad
\textsuperscript{6}Texas A\&M University\\
\textsuperscript{7}National University of Singapore\quad
\textsuperscript{8}Rice University\quad
\textsuperscript{9}Nanyang Technological University\par}
\vspace{4pt}
{\footnotesize\textsuperscript{\ensuremath{\ddagger}}Corresponding author:
\text{jiachen\_li@gatech.edu}\par}
}
}
\makeatletter
\IEEEaftertitletext{%
\begin{minipage}{\textwidth}
\centering\includegraphics[width=0.90\textwidth]{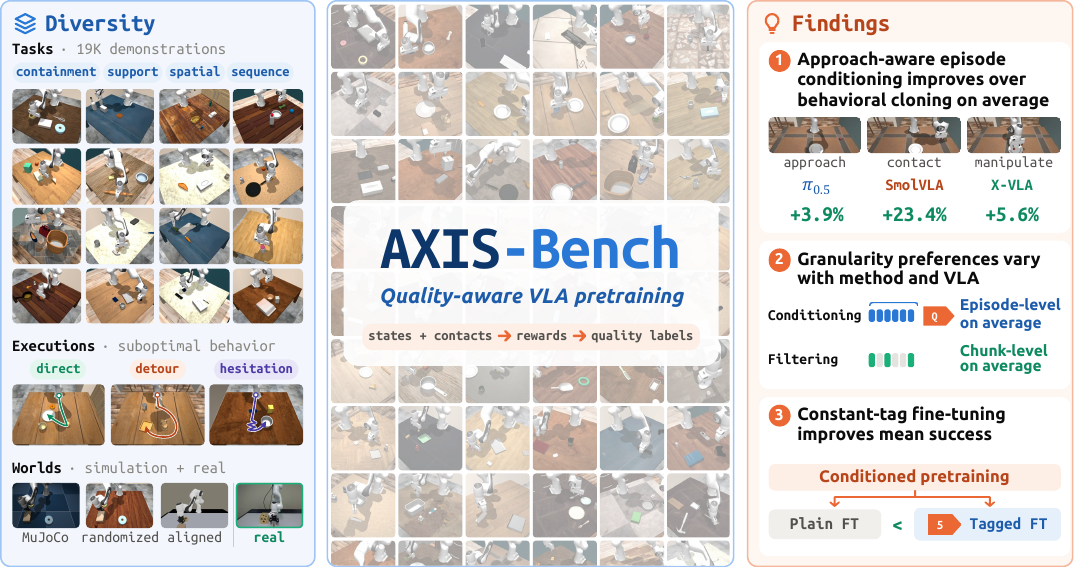}
\def\@captype{figure}
\caption{\textbf{AXIS-Bench enables a systematic study of quality-aware VLA pretraining.} We use 5,000 simulation demonstrations for controlled comparisons and 19,250 for pretraining before real-world fine-tuning. The simulation study spans three VLAs, comparing reward definitions, label granularities, and learning methods within each VLA. Green values show mean relative gains over plain behavioral cloning.
}
\label{fig:overview}
\end{minipage}%
\vspace{4pt}%
}
\makeatother

\begin{document}
\bstctlcite{axis:bstcontrol}

\maketitle
\thispagestyle{plain}\pagestyle{plain}\flushbottom

\begin{figure*}[!t]
\centering\includegraphics[width=\textwidth]{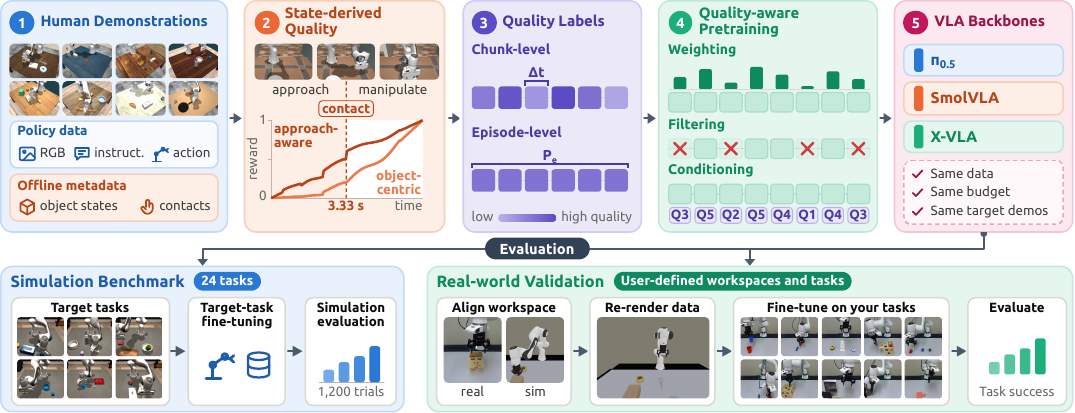}
\vspace{-0.6cm}
\caption{\textbf{AXIS-Bench workflow for quality-aware VLA pretraining.} Human demonstrations provide policy observations and action targets, while recorded object and contact states enable offline computation of state-derived quality. Quality is represented at the chunk or episode level and incorporated into pretraining through weighting, filtering, or conditioning across three VLA backbones.
The simulation benchmark evaluates performance after fine-tuning on 24 target tasks, with matched pretraining data, training budgets, and target-task demonstrations within each VLA comparison.
The real-world workflow supports user-defined workspaces and tasks by aligning the simulation workspace, re-rendering demonstrations under the target workspace and camera configuration, fine-tuning on real-world target-task demonstrations, and evaluating task success. Photos illustrate our reference real-world setup.
}
\label{fig:workflow}
\end{figure*}

\begin{abstract}
How should demonstration quality be used during vision-language-action (VLA) pretraining to improve performance after target-task fine-tuning? We compare weighting, filtering, and conditioning across reward definitions and label granularities using 5,000 naturally suboptimal human demonstrations and 24 target tasks. Within each VLA, we match demonstrations and training budgets. Additional comparisons examine pretraining data size and the roles of quality tags during fine-tuning and inference. Across three VLAs, we find that approach-aware episode conditioning improves mean success over plain behavioral cloning; conditioning and filtering favor episode and chunk labels, respectively, on average, with model-dependent reversals; and constant-tag fine-tuning improves mean success after conditioned pretraining with either tagged or untagged inference. We introduce AXIS-Bench to support these controlled comparisons. Built on crowdsourced simulation demonstrations, it provides recorded object and contact states for recomputing quality labels on shared trajectories. Replayable trajectories support workspace-aligned rendering and user-defined real-world target tasks. In our real-world experiment, simulation pretraining raises success from 9.6\% to 31.7\%. Project page: \url{https://axis-bench.github.io/}.
\end{abstract}

\vspace{-18pt}
\section{Introduction}
Vision-language-action (VLA) policies are commonly pretrained on multitask demonstrations and fine-tuned on target tasks before deployment \cite{o2024open,team2024octo,pmlr-v270-kim25c,pmlr-v305-black25a}. During a demonstration, an operator may hesitate or move an object along an unnecessary detour, producing action chunks that make little progress toward the goal. Quality-aware pretraining~\cite{pmlr-v270-hejna25a,intelligence2026pi} aims to learn effectively from naturally suboptimal demonstrations.
We evaluate its effect on performance after target-task fine-tuning (FT).

Quality-based filtering, weighting, and conditioning have been studied in robot learning \cite{hejna2025robot,pmlr-v270-hejna25a,zhang2026scizor,intelligence2025pi,intelligence2026pi}. Yet a quality signal leaves several learning choices open. The same quality information can be used to select which training examples to retain~\cite{zhang2026scizor,wang2026world}, determine how strongly individual actions contribute to the training loss~\cite{peng2019advantage,wang2026world}, or condition the policy through explicit quality tags~\cite{intelligence2026pi}.
Quality can also be defined at different granularities, describing local progress over an action chunk or summarizing the quality of an entire episode. For conditioned policies, additional choices arise regarding how quality tags should be handled during subsequent fine-tuning and inference.
However, systematic evidence comparing how these choices interact under shared data and evaluation protocols remains limited.
To address this gap, we study how quality definitions, label granularities, and learning methods jointly affect target-task performance, and which findings are consistent across base VLA policies. We compare weighting, filtering, and conditioning across two progress-based reward definitions, two label granularities, and three VLAs ($\pi_{0.5}$~\cite{pmlr-v305-black25a}, SmolVLA~\cite{shukor2025smolvla}, and X-VLA~\cite{zheng2026x}). Using 5,000 pretraining demonstrations and 24 target tasks, we further examine the effects of pretraining data size and fine-tuning/inference protocols.

Three findings emerge (see Fig.~\ref{fig:overview}). First, approach-aware episode conditioning improves mean success over plain behavioral cloning (plain BC) across the three evaluated VLAs. Second, label granularity interacts with both the learning method and base VLA: conditioning favors episode labels and filtering favors chunk labels on average, but the preferences can reverse. Third, after approach-aware episode conditioning, constant-tag target-task FT yields higher mean success than plain FT with tagged/untagged inference.

To support this study, we introduce AXIS-Bench (see Fig.~\ref{fig:workflow}), a simulation and real-world benchmark built on the crowdsourced AXIS simulation teleoperation dataset \cite{zhao2026axis}.
Recorded object and contact states let us recompute progress-based quality labels on the same trajectories.
Replayable trajectories further enable researchers to re-render demonstrations to align with target workspace and camera configurations and evaluate policies on their own real-world target tasks.
In our real-world reference experiment, simulation pretraining raises $\pi_{0.5}$-base success rate from 9.6\% to 31.7\%.

Our main contributions are:
\begin{enumerate}
\item \textbf{A systematic empirical study of quality-aware VLA pretraining.} We identify how quality definitions, label granularities, learning methods, and adaptation protocols interact across base policies, with matched data and training budgets within each VLA.
\item \textbf{AXIS-Bench for controlled comparison and real-world validation.} We provide shared human demonstrations, configurable state-derived quality labels, and standardized training and evaluation protocols, together with replayable simulation trajectories and a real-world setup for workspace adaptation.
\end{enumerate}

\section{Related Work}
\textbf{Robot demonstration datasets.} Robot demonstration datasets have expanded the diversity of tasks, environments, and embodiments available for policy learning. BridgeData V2 and DROID collect real-world demonstrations across tasks and workspaces \cite{pmlr-v229-walke23a,khazatsky2024droid}, while Open X-Embodiment aggregates data across robot embodiments \cite{o2024open}, supporting generalist policies such as Octo and OpenVLA \cite{team2024octo,pmlr-v270-kim25c}.
Simulation provides complementary approaches through crowdsourced teleoperation, as in RoboTurk \cite{pmlr-v87-mandlekar18a}, and demonstration-based or procedural generation, as in MimicGen and MolmoB0T \cite{pmlr-v229-mandlekar23a,deshpande2026molmob0t}.
Real-world collections naturally capture variation in human execution but generally lack simulator states for directly recomputing progress-based rewards. Generated simulation trajectories provide privileged state information but need not preserve the natural execution variability of human demonstrations. AXIS unifies crowdsourced human teleoperation with recorded object states, contacts, and replayable trajectories \cite{zhao2026axis}.

\textbf{Quality-aware VLA training.}
Studies of demonstration proficiency motivate learning selectively from demonstrations of varying quality \cite{pmlr-v164-mandlekar22a}. 
Related data-curation methods estimate training utility using state-action mutual information, target-task relevance, or rollout-based influence \cite{hejna2025robot,dass2026datamil,pmlr-v305-agia25a,ghosh2024robustofflineimitationlearning}. These methods also operate at different granularities: Re-Mix optimizes dataset-level mixture weights \cite{pmlr-v270-hejna25a}, SCIZOR filters individual transitions for generalist policy training \cite{zhang2026scizor}, and World Value Models (WVMs) use chunk-level value changes for weighting and filtering \cite{wang2026world}.
Conditioning supplies quality information as an input to the policy. Decision Transformer and RvS study return-conditioned supervised learning for offline RL \cite{3540261.3541417,emmons2022rvs}. In VLA training, RECAP applies advantage conditioning during both pretraining and task-specific training \cite{intelligence2025pi}, while $\pi_{0.7}$ conditions on quality, speed, and mistake metadata \cite{intelligence2026pi}. Related metadata- and quality-conditioning strategies have also been explored in language-model pretraining and vision-language learning \cite{korbak2023pretraining,3780338.3781058,kang2023noise}.
We complement these studies with a controlled comparison of reward definitions, label granularities, and learning methods on shared demonstrations across three VLAs.

\section{Benchmark Design and Data Resource}

\subsection{Human demonstrations and replayable data}
AXIS-Bench pairs human demonstrations with scene assets and replayable simulation trajectories (see Fig.~\ref{fig:workflow}). Policy observations comprise images, task instructions, and proprioceptive states, paired with corresponding action targets. Recorded object geometry, contacts, task progress, episode boundaries, and validity masks support offline quality labeling (Sec.~\ref{sec:quality}). We use two pretraining datasets: 5,000 simulation demonstrations (about 35.3 hours) for the simulation experiments and 19,250 simulation demonstrations (about 114.6 hours) for the real-world experiments.
Both simulation and real-world experiments use a DROID-compatible interface \cite{khazatsky2024droid}, with actions comprising seven joint velocities and a gripper command at 15\,Hz. Pretraining observations include both external-camera and wrist-camera views.

\subsection{Controlled simulation benchmark}
The simulation branch of Fig.~\ref{fig:workflow} evaluates target-task performance on 24 tabletop tasks spanning containment, support placement, spatial relations, separation, and short sequences. Pretraining uses the shared 5,000-demonstration dataset, in which object meshes and poses, language goals, and visual conditions vary across demonstrations. Visual randomization varies room backgrounds, table textures, and lighting while keeping the external-camera pose fixed. Target-task FT uses 354 demonstrations (about 2.4 hours). The embodiment, renderer, and object library are shared between pretraining and evaluation, so the benchmark evaluates target tasks and initial object placements within this setting. Pretraining and target task IDs are disjoint, although three target instructions also occur under four pretraining task IDs (22 of the 5,000 demonstrations); the split therefore does not imply complete semantic separation.
Within each VLA configuration, method comparisons use the same base-policy initialization, training budget, target-task demonstrations, and evaluation layouts.

\subsection{Workspace alignment and user-defined target tasks}
For real-world validation (Fig.~\ref{fig:workflow}), researchers can align the simulated tabletop, background, and camera poses with their target workspace. Re-rendering preserves the recorded actions and object states while producing pretraining observations aligned with the target workspace and camera configuration. Researchers then collect real-world demonstrations for target-task FT and evaluate policies under fixed task definitions and success criteria. Our reference implementation uses robosuite \cite{zhu2020robosuite}, built on MuJoCo \cite{6386109}, to re-render trajectories for a Franka FR3 workspace.

\section{State-Derived Quality and Pretraining}
\label{sec:quality}
We use progress toward the task goal as a measure of demonstration quality for learning from naturally suboptimal human data. As illustrated in Fig.~\ref{fig:workflow}, we compute progress-based rewards from recorded states, derive quality labels at the chunk or episode level, and incorporate them into pretraining through weighting, filtering, or conditioning.

\subsection{State-derived progress rewards}

\begin{figure}[t]
\centering\includegraphics[width=\linewidth]{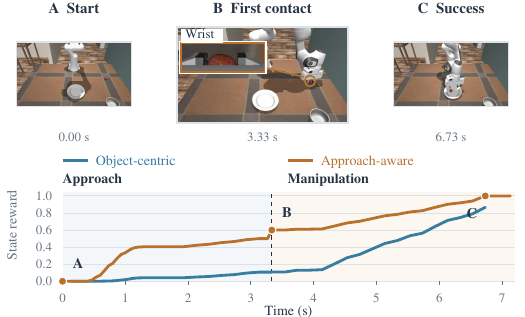}
\vspace{-0.3cm}
\caption{\textbf{Two state-derived reward definitions.} A, B, and C denote the start, first contact, and task success, respectively. B includes the corresponding wrist-camera view. First sustained contact marks the transition from approach to manipulation for the approach-aware reward.}
\label{fig:reward}
\end{figure}

We derive progress signals from simulator trajectories to quantify how each demonstration advances toward the task goal (Fig.~\ref{fig:reward}). Task-specific object progress can remain unchanged during both purposeful approach motion and hesitation. We compare a reward that combines object progress with end-effector proximity and an approach-aware reward that treats approach and manipulation separately. Let $d_t$ denote the distance between the end effector and the manipulated object, and $p_t\in[0,1]$ the task-specific object progress at time step $t$. The \textit{object-centric reward} is
\vspace{-0.2cm}
\begin{equation}
    r_t=0.8p_t+0.2c_t,
\end{equation}
where $c_t=1-\tanh(5d_t)$ captures end-effector proximity. For the approach-aware reward, we isolate the task-progress component as \begin{equation}p_t=\clip((r_t-0.2c_t)/0.8,0,1).
\end{equation}

The approach-aware signal uses the first sustained finger-object contact as a task-agnostic boundary between approach and manipulation. Before contact, progress is measured by the fraction of the required approach completed. After contact, it is measured by the additional object-centric progress achieved relative to the contact state. Let $t_c$ denote the first frame in a sequence of at least three consecutive finger-object contact frames. We define the \textit{approach-aware reward} as
\begin{equation} R_t= \begin{cases} \lambda\clip\left(\dfrac{d_0-d_t}{d_0-d_{t_c}},0,1\right), & t<t_c,\\[5pt] \lambda+b+(1-\lambda-b)\clip(p_t-p_{t_c},0,1), & t\ge t_c, \end{cases} \label{eq:phase} \end{equation}
with $\lambda=0.5$ and $b=0.1$.
This construction scores approach and manipulation separately. Episodes with contact at the start or no measurable approach use the manipulation branch from the start; episodes with no contact or contact only at the tail use approach progress normalized by the closest approach reached. Episodes without a valid object assignment remain unscored.

\subsection{From reward changes to quality labels}

We convert each state-derived reward into chunk-level and episode-level quality scores.
For an episode of $T$ time steps and either reward $z_t\in\{r_t,R_t\}$, we first compute the raw reward change over a fixed $H=10$ frame window:
\begin{equation}
\Delta_t^{\mathrm{raw}}
=
z_{\min(t+H-1,T-1)}-z_t,
\qquad H=10.
\label{eq:chunk}
\end{equation}
We obtain the final chunk quality score $\Delta_t$ by applying validity and phase-boundary rules to this raw change. For the approach-aware reward, windows that cross the contact boundary are truncated so that the contact offset is not counted as local progress. If $h_t$ intervals remain after truncation, we rescale the within-phase change by $(H-1)/h_t$; windows with $h_t<3$ or invalid reward values receive a score of zero. Positive, zero, and negative values of $\Delta_t$ indicate progress, no measured progress, and regression, respectively.

The \textit{episode quality score} is the fraction of retained time steps whose corresponding action chunks make positive progress:
\begin{equation}
P_e=
\frac{1}{|\mathcal T_e|}
\sum_{t\in\mathcal T_e}
\mathbf{1}[\Delta_t>0],
\label{eq:episode}
\end{equation}
where $\mathcal T_e$ contains the retained time steps from episode $e$. Thus, $\Delta_t$ captures the magnitude of local progress, whereas $P_e$ captures how consistently an execution makes progress over the episode.

For conditioning, we rank valid quality scores across the pretraining dataset and partition them into five equal-frequency bins, yielding labels $q\in\{1,\ldots,5\}$ from low to high quality. Chunk-level labels are derived from $\Delta_t$, whereas episode-level labels are derived from $P_e$ and shared across all retained time steps within an episode.

\begin{figure}[!t]
\centering\includegraphics[width=\linewidth]{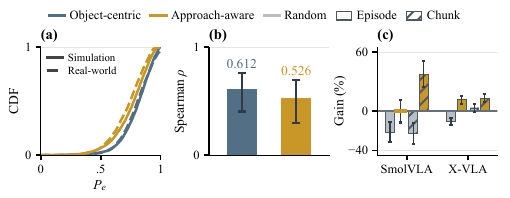}
\vspace{-0.8cm}
\caption{\textbf{Quality distributions, alignment with human judgments, and utility for data selection.} (a) CDFs of episode-level quality scores for all 5,000 simulation episodes and 17,758 scorable simulation episodes from the 19,250 used for real-world experiments; object-centric scores for the latter are shown for reference. (b) Spearman correlations between state-derived episode scores and human judgments, with 95\% episode-bootstrap confidence intervals (10,000 resamples). (c) Random selection versus approach-aware quality filtering at the same half-data budget for SmolVLA and X-VLA; solid and hatched bars denote episode- and chunk-level selection, respectively. Gains are relative to each VLA's full-data plain-BC baseline; whiskers show rescaled marginal $\pm1$ SE.}
\label{fig:quality-validation}
\end{figure}

We further assess whether the state-derived episode scores align with human judgments. Three annotators mark unproductive intervals in 50 simulation demonstrations, and we define human-assessed episode quality as
$H_e=1-\operatorname{median}_j U_{e,j}$,
where $U_{e,j}$ is the fraction of judgeable retained time steps marked unproductive by annotator $j$. Ambiguous annotations are excluded, leaving at least two annotators per episode. \textit{The state-derived episode scores correlate positively with human judgments}, with Spearman $\rho(P_e,H_e)=0.612$ for the object-centric reward and $0.526$ for the approach-aware reward (Fig.~\ref{fig:quality-validation}(b)).
To assess their utility for policy learning, we compare quality filtering with random selection at each granularity, matching retained time steps and training updates (Fig.~\ref{fig:quality-validation}(c)). For SmolVLA and X-VLA, approach-aware filtering improves over random selection by 1.08 and 5.75 percentage points (pp) at episode level, and by 3.08 and 2.58 pp at chunk level, respectively. The benefit depends on reward choice: object-centric chunk filtering exceeds random selection by 3.17 pp for SmolVLA but falls below it by 3.00 pp for X-VLA.

\begin{figure*}[t]
\centering\includegraphics[width=\textwidth]{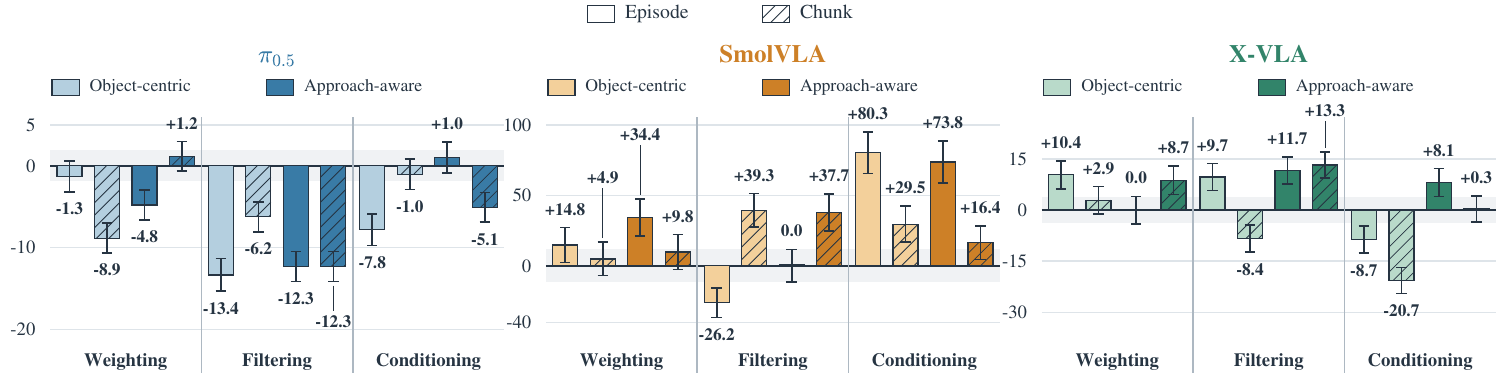}
\caption{\textbf{Quality-aware learning on 24 tasks}: AWR-style weighting, top-50\% filtering, and conditioning. Light/dark shades indicate object-centric/approach-aware rewards; solid/hatched bars indicate episode/chunk labels. Values are relative improvements (\%) over plain BC, whose success rates are 57.42\%, 5.08\%, and 25.75\% for $\pi_{0.5}$, SmolVLA, and X-VLA, respectively. Matched FT-only controls reach 8.83\%, 2.50\%, and 6.17\%. Whiskers/bands show rescaled marginal $\pm1$ SE.}
\label{fig:granularity}
\end{figure*}

\subsection{Weighting, filtering, and conditioning}

We compare three ways of incorporating the same state-derived quality scores into pretraining: loss weighting, sample filtering, and quality-label conditioning. Chunk-level methods use $\Delta_t$, whereas episode-level methods use $P_e$.

\textbf{Plain BC and weighting.}
Plain BC samples retained demonstration time steps uniformly and optimizes the standard behavioral cloning loss. For quality weighting, following advantage-weighted regression (AWR)~\cite{peng2019advantage} and WVM~\cite{wang2026world}, we assign each time step the chunk-level weight
$w_t=\min(\exp(10\Delta_t),2)$
and optimize the weighted loss
$\sum_t w_t\ell_t/\sum_t w_t$,
where $\ell_t$ is the flow-matching loss for the action chunk starting at time step $t$. For episode-level weighting, all time steps within an episode share a weight determined by its $P_e$. To maintain a comparable weight distribution across granularities, we map the time step-weighted percentile rank of $P_e$ to the empirical distribution of chunk-level weights.

\textbf{Filtering.}
Top-quality-50\% filtering retains half of the preprocessed training time steps. Chunk-level filtering ranks time steps by $\Delta_t$, whereas episode-level filtering ranks episodes by $P_e$ and retains the highest-quality episodes until the same time step budget is reached. When necessary, the boundary episode is partially retained to match the budget exactly. We repeat the selected data to match the number of training updates used by plain BC.

\textbf{Conditioning.}
Quality conditioning retains all training data with uniform sampling and appends a \texttt{Quality: $q$} tag to each language instruction following $\pi_{0.7}$~\cite{intelligence2026pi}. In simulation, tagged target-task FT assigns tag~5 to every demonstration before tag dropout, and conditioned inference also uses tag~5. We apply quality-tag dropout during pretraining and tagged FT; the dropout settings and real-world conditioning protocol appear in the Appendix.

\section{Simulation Experiments}
We ask three questions: which quality-aware pretraining choices improve performance across VLAs, how label granularity interacts with the learning method, and whether quality tags should be retained during target-task FT. We also examine how performance changes with dataset size.

\subsection{Experimental overview}
\begin{figure}[t]
\centering\includegraphics[width=\linewidth]{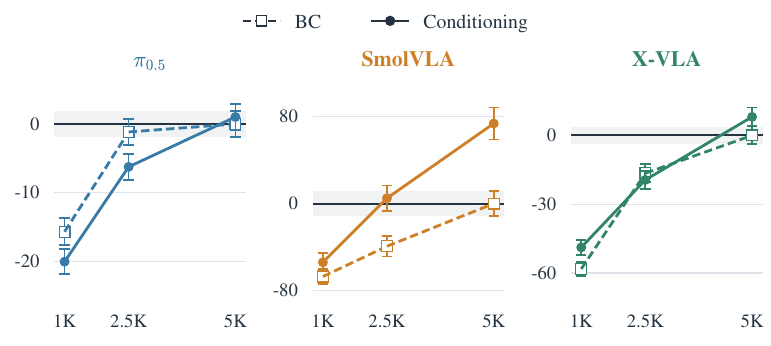}
\caption{\textbf{Scaling with pretraining data size.} Values are relative improvements (\%) over each VLA's fixed 5K plain BC. Dashed squares denote plain BC with plain FT/untagged inference; solid circles denote conditioning with tagged FT/tag~5. Whiskers show rescaled marginal SE; shaded bands show plain BC uncertainty.}
\label{fig:scale}
\end{figure}

\begin{figure*}[!t]
\centering\includegraphics[width=\textwidth]{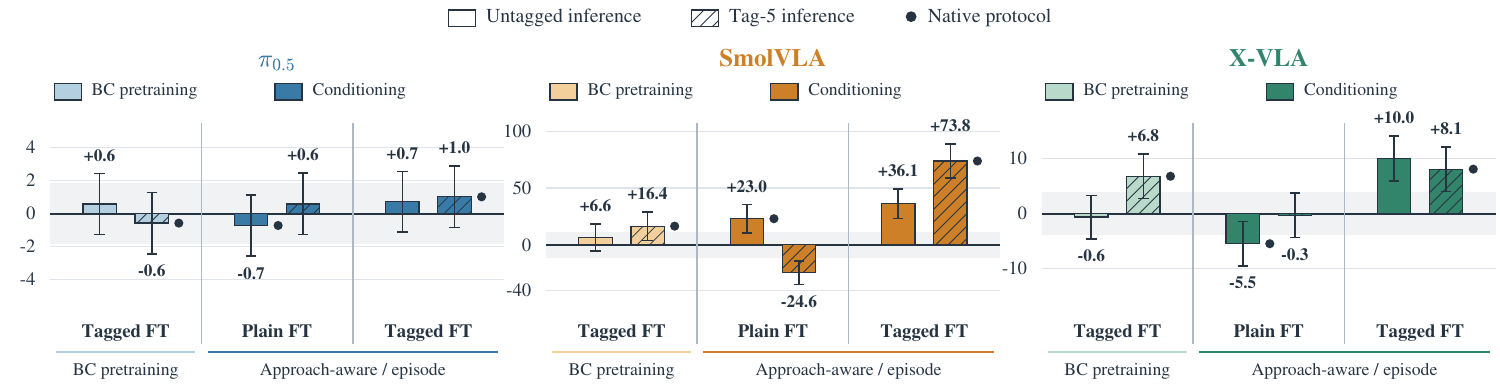}
\caption{\textbf{Conditioning stage and inference context.} Light/dark shades indicate plain BC/conditioned pretraining; solid/hatched bars indicate untagged/tag-5 inference. $\bullet$ marks native protocols: untagged after plain FT, tag~5 after tagged FT. Values are relative improvements (\%) over plain BC/plain-FT/untagged inference. Whiskers/bands show rescaled marginal SE.}
\label{fig:attribution}
\end{figure*}

We compare plain BC, weighting, filtering, and conditioning on 5,000 pretraining demonstrations, followed by FT on 354 demonstrations covering 24 target tasks. Within each VLA, comparisons share the initialization, data, training budget, and evaluation layouts. We initialize $\pi_{0.5}$~\cite{pmlr-v305-black25a} from the released $\pi_{0.5}$-DROID checkpoint and X-VLA~\cite{zheng2026x} from its released base policy; SmolVLA~\cite{shukor2025smolvla} uses a frozen pretrained VLM and a randomly initialized action expert. Plain BC, weighting, and filtering use plain FT and untagged inference; conditioning uses constant-tag FT and tag-5 inference, with stage-specific controls below. Matched FT-only controls omit AXIS pretraining.
For each of the 24 tasks, we evaluate all configurations on the same set of 50 initial scene layouts (1,200 trials per configuration) and report mean task success. Error bars in Figs.~\ref{fig:granularity}--\ref{fig:attribution} denote $\pm1$ evaluation SE. Results are plotted relative to each VLA's plain-BC reference, with absolute success rates given in the figure captions. Detailed training, evaluation, and plotting protocols are in appendix.

\subsection{Cross-VLA gains from episode conditioning}
\textbf{Finding 1: Approach-aware episode conditioning improves mean success across all three VLAs.}
Fig.~\ref{fig:granularity} compares all 36 combinations of VLA, reward definition, label granularity, and learning method. With tagged FT and tag-5 inference, approach-aware episode conditioning improves over plain BC by 1.02\%, 73.77\%, and 8.09\% relative for $\pi_{0.5}$, SmolVLA, and X-VLA, respectively.
Method rankings vary across models. Across the four reward-granularity settings, $\pi_{0.5}$ conditioning and weighting have similar means above filtering; SmolVLA favors conditioning, whereas X-VLA favors filtering and weighting. X-VLA filtering exceeds conditioning in all four matched settings.

\subsection{Coupling between label granularity and learning method}
\label{sec:quality-results}
\textbf{Finding 2: Conditioning favors episode labels and filtering favors chunk labels on average, with VLA- and reward-dependent exceptions.} In Fig.~\ref{fig:granularity}, episode conditioning exceeds chunk conditioning in five of six matched VLA--reward settings. Chunk filtering exceeds episode filtering in four settings, ties in one, and falls below it in one. Under approach-aware reward, episode conditioning exceeds chunk conditioning by 3.50, 2.92, and 2.00 pp for $\pi_{0.5}$, SmolVLA, and X-VLA, respectively, while chunk filtering matches or exceeds episode filtering for each VLA. The two reversals occur with object-centric reward: $\pi_{0.5}$ conditioning favors chunk labels by 3.92 pp, and X-VLA filtering favors episode labels by 4.67 pp. An episode label may help conditioning distinguish overall execution patterns through a consistent quality context. Chunk filtering instead selects locally productive training start points, including those within episodes containing unproductive behavior. For X-VLA, random chunk selection also exceeds random episode selection by 3.58 pp (Fig.~\ref{fig:quality-validation}(c)), suggesting that data composition may contribute to its granularity preference.

Weighting has reward-dependent preferences: object-centric weighting favors episode labels across all three models; approach-aware weighting favors chunk labels for $\pi_{0.5}$ and X-VLA, but episode labels for SmolVLA. Approach-aware chunk weighting also improves mean success over BC across models, with small gains for $\pi_{0.5}$ and SmolVLA (+0.67 and +0.50 pp).

\subsection{Retaining quality tags during fine-tuning}
\textbf{Finding 3: After conditioned pretraining, tagged FT improves mean success under both inference modes.} Fig.~\ref{fig:attribution} compares plain and constant-tag FT after approach-aware episode conditioning under both inference modes for all three VLAs. We compare plain and constant-tag FT within each VLA, keeping the inference protocol fixed. Inference comparisons evaluate each FT checkpoint with and without tag~5.
With tag-5 inference, tagged FT exceeds plain FT by 0.25, 5.00, and 2.17 pp for $\pi_{0.5}$, SmolVLA, and X-VLA, respectively. Under untagged inference, the corresponding differences are 0.83, 0.67, and 4.00 pp. The $\pi_{0.5}$ differences are small; SmolVLA gains more with tag-5 inference, and X-VLA without it.
Adding tag~5 at inference changes SmolVLA success by $-2.42$ pp after plain FT and +1.92 pp after tagged FT. X-VLA shows the opposite pattern (+1.33 and $-0.50$ pp, respectively), illustrating dependence on the VLA and FT protocol. After plain-BC pretraining, switching from plain FT/untagged inference to tagged FT/tag-5 inference has mixed effects: $-0.33$, +0.83, and +1.75 pp for $\pi_{0.5}$, SmolVLA, and X-VLA. Adding the tag only at inference also has opposite effects for $\pi_{0.5}$ and SmolVLA (+0.92 and $-0.67$ pp).
The constant FT tag supplies no quality ranking among target-task demonstrations. Retaining the pretraining input format may instead help preserve the policy's response to conditioning during adaptation. Studies of metadata conditioning~\cite{3780338.3781058}, prompt templates~\cite{lyu2024keeping}, and implicit inference after FT~\cite{kotha2024understanding} motivate this hypothesis. These opposing inference effects show that tag~5 alone is not always beneficial.

\textit{Overall Finding: Taken together, these results highlight the need to jointly consider reward definition, label granularity, learning method, and target-task adaptation. No single pretraining configuration achieves the highest success rate across all evaluated VLAs.}

\subsection{Scaling with dataset size}
Fig.~\ref{fig:scale} compares conditioning with plain BC at matched budgets of 1K, 2.5K, and 5K demonstrations, plotted relative to each VLA's fixed 5K BC reference. For $\pi_{0.5}$, conditioning falls below BC at the smaller budgets and slightly exceeds it at 5K (58.00\% versus 57.42\%). SmolVLA conditioning exceeds BC at all three sizes, reaching 8.83\% versus 5.08\% at 5K. X-VLA conditioning exceeds BC at 1K and 5K, but falls below it at 2.5K (20.75\% versus 21.50\%). \textbf{Larger pretraining datasets improve success in these sweeps, while the conditioning advantage depends on the VLA and dataset size.} The sweep scales training updates with data size, so these results reflect the combined change in data and computation.

\subsection{Training-seed replication}
\label{sec:seed-replication}
To assess the consistency of the findings across training runs, we evaluate the approach-aware comparisons with three training seeds per VLA: 42/43/44 for $\pi_{0.5}$ and 0/1/2 for SmolVLA and X-VLA. The method, scaling, and FT/inference comparisons in Figs.~\ref{fig:granularity}--\ref{fig:attribution} use seed 42 for $\pi_{0.5}$ and seed 0 for SmolVLA and X-VLA. We use the 24-task evaluation protocol above, with the same 50 layouts for a given task across all configurations. Table~\ref{tab:seed-replication} summarizes the within-seed success-rate differences normalized by the corresponding plain-BC success rate.
\begin{table}[t]
\caption[Training-seed replication]{\parbox[t]{\columnwidth}{\normalfont\footnotesize\textbf{Training-seed replication.} Entries are mean $\pm$ sample SD of BC-normalized success-rate differences (\%) over three seeds per VLA, using approach-aware reward. For each row $A-B$, we compute $100(\mathrm{SR}_A-\mathrm{SR}_B)/\mathrm{SR}_{\mathrm{BC}}$ within each seed before averaging. Episode conditioning uses tagged FT/tag-5 inference; BC uses plain FT/untagged inference. The last two rows compare tagged and plain FT after episode-conditioned pretraining, with inference fixed as indicated.}}
\label{tab:seed-replication}
\centering
\setlength{\tabcolsep}{2pt}
\renewcommand{\arraystretch}{1.15}
\fontsize{7.5}{9}\selectfont
\begin{tabular*}{\columnwidth}{@{\extracolsep{\fill}}lccc@{}}
\hline
Comparison & $\pi_{0.5}$ & SmolVLA & X-VLA \\
\hline
Episode cond. $-$ BC & $+3.86\pm5.83$ & $+23.42\pm45.42$ & $+5.62\pm5.39$ \\
Cond.: episode $-$ chunk & $-0.90\pm8.52$ & $+24.71\pm32.64$ & $+11.36\pm15.16$ \\
Filter: chunk $-$ episode & $-1.90\pm3.56$ & $+6.63\pm28.34$ & $+5.59\pm8.05$ \\
Tagged $-$ plain: untagged & $+0.36\pm4.30$ & $+9.58\pm12.31$ & $+7.86\pm6.81$ \\
Tagged $-$ plain: tag 5 & $+0.93\pm4.28$ & $+62.29\pm31.29$ & $+8.93\pm1.95$ \\
\hline
\end{tabular*}
\end{table}
\textbf{The three-seed results retain positive mean gains from episode conditioning and tagged FT. Averaged across VLAs and seeds, conditioning favors episode labels and filtering favors chunk labels, with model- and seed-dependent reversals.} Conditioning exceeds plain BC in six of nine seed-level comparisons, with one tie and two negative differences.
On this BC-normalized scale, equal averaging over all nine VLA--seed combinations favors episode labels for conditioning ($+11.72\%$) and chunk labels for filtering ($+3.44\%$). Both directions hold in the SmolVLA and X-VLA means but reverse for $\pi_{0.5}$.

All six VLA--inference comparisons favor tagged FT on average, with 15 of 18 seed-level differences positive. Inference-time tags show a different pattern: after tagged FT, removing the inference tag improves X-VLA success at all three seeds, while adding it improves SmolVLA success in one run and reduces it in two.

\section{Real-World Experiments}
\label{sec:real}
\subsection{Experimental overview}
We evaluate simulation pretraining on ten real-world manipulation tasks in a Franka FR3 workspace, with 25 real-world demonstrations collected per task for target-task fine-tuning. Each policy is evaluated with 24 trials per task (240 trials in total). All three configurations start from $\pi_{0.5}$-base and use the same fine-tuning dataset. We compare an FT-only control with plain BC and approach-aware episode conditioning pretrained on 19,250 simulation demonstrations. The FT-only and plain-BC configurations use plain FT and untagged inference; conditioning uses tagged FT and tag-5 inference. Hardware, training, and evaluation details appear in the appendix.

Fig.~\ref{fig:workspace-alignment} illustrates workspace alignment: we first adjust the simulated camera view, then configure the tabletop and background to match the real workspace. Re-rendering changes the observations while preserving recorded actions and object states. Configuring our reference workspace took approximately 30 minutes, after which recorded trajectories could be re-rendered for pretraining.

\begin{figure}[!t]
\centering\includegraphics[width=\linewidth]{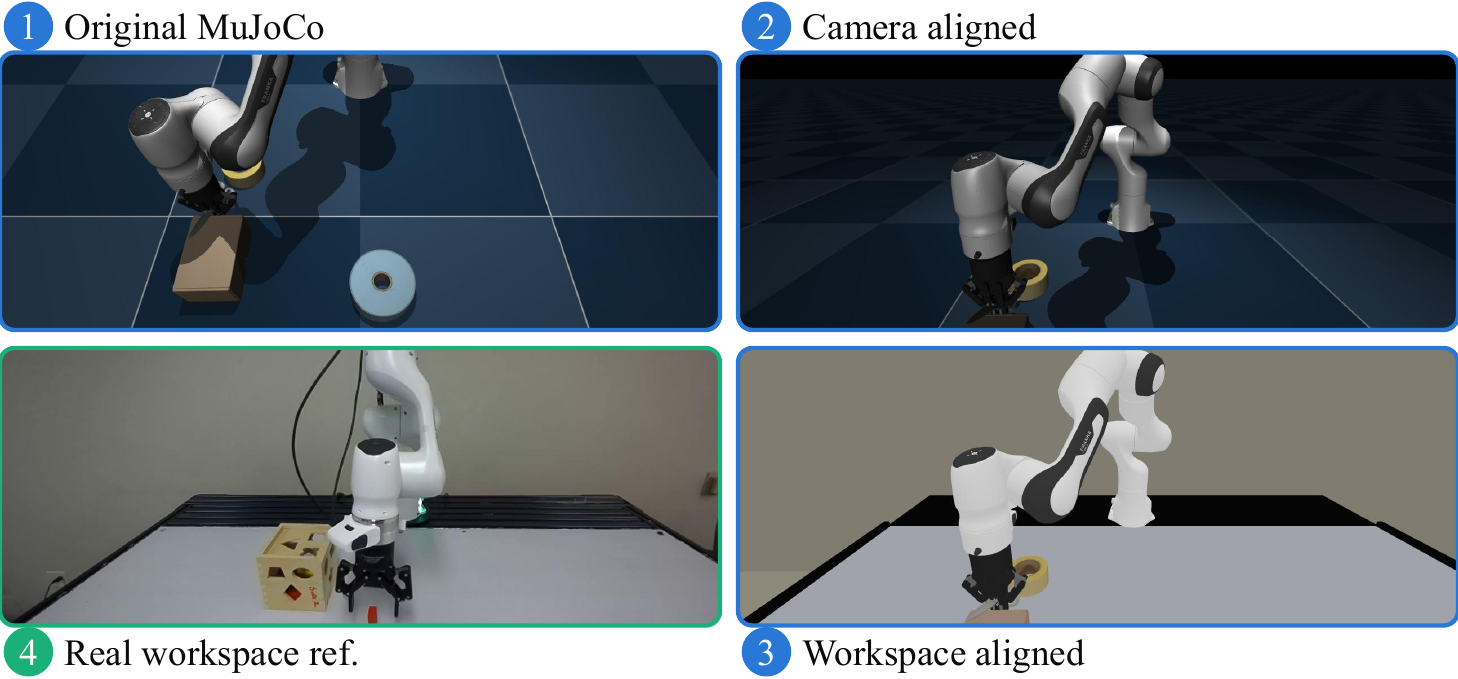}
\caption{\textbf{Workspace alignment for real-world pretraining.} The original MuJoCo scene (1) is re-rendered with an aligned camera view (2), followed by tabletop and background alignment (3). The real workspace (4) provides the visual reference. Recorded actions and object states are preserved during re-rendering.}
\label{fig:workspace-alignment}
\end{figure}

\subsection{A simulation-pretraining baseline for real-world tasks}
The plain-BC and no-pretraining arms differ only in whether AXIS pretraining precedes the same plain FT and untagged inference. Success rises from 9.6\% to 31.7\%, improving on nine of ten tasks. \textbf{Simulation~pretraining improves real-world target-task success}, establishing a baseline for quality-aware methods.
\begin{figure}[!tbp]
\centering\includegraphics[width=\linewidth]{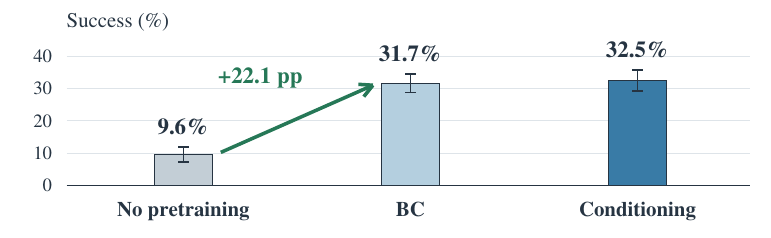}
\vspace{-20pt}
\caption{\textbf{Real-world improvement from simulation pretraining.} All policies start from $\pi_{0.5}$-base. The FT-only reference and plain BC use plain FT/untagged inference; conditioning uses tagged FT/tag~5. Whiskers show $\pm1$ evaluation SE over layout groups.}
\label{fig:real}
\end{figure}
Conditioning achieves 32.5\% success, a gain of 0.83 pp over plain BC, with five tasks improving, four decreasing, and one tied. This real-world evaluation uses a different initialization, data size, and training recipe from the simulation benchmark.

\raggedbottom
\section{Discussion and Conclusion}
In the evaluated configurations, approach-aware episode conditioning improves mean success over plain BC, and constant-tag FT improves mean success after conditioned pretraining under both tested inference protocols. Conditioning favors episode labels and filtering favors chunk labels on average, with model-dependent reversals.
The study covers a shared tabletop simulation setting, with real-world validation using one VLA in a reference workspace. Testing larger pretraining budgets and additional workspaces and embodiments would clarify how broadly the findings transfer. Initialization and trainable modules vary across VLAs, so the comparisons do not isolate the causes of their different method rankings. Matched controls could test whether retaining data coverage explains conditioning's benefits, and which aspects of initialization and adaptation explain the observed differences.

Our progress-based labels capture only part of demonstration quality and may undervalue necessary preparation or recovery with little immediate progress. Preference-based reward learning~\cite{ghosh2026reducing} and personalized policy conditioning~\cite{wang2026drive} suggest ways to incorporate user-dependent objectives beyond task progress. Demonstration decomposition for planner alignment~\cite{yan2025rdd} also motivates examining quality over meaningful subtasks. Whether such signals improve quality-aware pretraining remains an open question.

Quality supervision could also complement task-relevant visual attention~\cite{gong2026autofocus} and semantic guidance for pretrained policies~\cite{gao2025guides}. Beyond nominal task success, deployment requires evaluating robustness to sensory and instruction perturbations~\cite{jiang2026navtrust} and scene--task inconsistencies~\cite{chakraborty2025heal}. Safety alignment through constrained learning~\cite{zhang2025safevla} addresses a further objective that progress-based quality labels do not explicitly encode. These directions would extend the present evaluation beyond task completion to the reliability and safety of learned behavior.
AXIS-Bench supports further comparisons by enabling new quality labels to be computed and trajectories to be re-rendered for user-defined workspaces and tasks.

\clearpage
\flushbottom
\appendix[Experimental Protocols]
\subsection{Simulation experiment details}
\label{app:sim-training}
\label{app:sim-evaluation}
For $\pi_{0.5}$, pretraining updates all parameters for 20,605 updates (batch 64, learning rate $5\times10^{-5}$). FT starts from the EMA checkpoint (decay 0.999) and runs for 13,580 updates (batch 32, learning rate $2.5\times10^{-5}$), using LoRA for language/action modules and a trainable vision encoder. Pretraining and FT use 2,060 and 500 warmup updates, respectively. Target demonstrations follow a fixed within-task ranking by approach-aware progress established during dataset curation.
SmolVLA and X-VLA use 20,621 pretraining and 6,790 FT updates, with batch 64 and learning rate $10^{-4}$. SmolVLA starts from a pretrained SmolVLM2 with a randomly initialized action expert and updates only the action expert and state projection, keeping the VLM and vision encoder frozen; X-VLA updates all parameters, with one-tenth the learning rate for its VLM. Simulation normalization statistics are fitted separately to each stage's data and shared across methods. All models receive five FT epochs, and we evaluate the final checkpoint. Images use $224\times224$ center crops for $\pi_{0.5}$ and padding to $512\times512$ for SmolVLA or $224\times224$ for X-VLA. SmolVLA uses AdamW with cosine scheduling; X-VLA uses domain-prompt index 20.

Random-selection controls use a fixed sampling seed and match the retained time steps and training updates of quality filtering.
Quality-tag dropout in simulation uses two independent dropout operations with probabilities $0.15$ and $0.05$ during both pretraining and tagged FT. Pretraining masks are sampled once offline and shared across corresponding experiments; tagged-FT masks are resampled each epoch.
For evaluation, the 24 tasks comprise six containment, six support-placement, seven spatial-relation, two separation, and three short-sequence tasks. The native prediction horizons are 10, 50, and 30 actions for $\pi_{0.5}$, SmolVLA, and X-VLA, respectively. Policies return the first ten predicted actions, and the client executes eight before replanning. The $H=10$ labeling window is independent of the native prediction horizon. Matched FT-only controls omit AXIS pretraining while retaining the corresponding initialization.

Relative improvements use $100(\mathrm{SR}/\mathrm{SR}_{\mathrm{BC}}-1)$ with unrounded success rates; marginal SEs are scaled by $100/\mathrm{SR}_{\mathrm{BC}}$. The reference is each VLA's 5K plain BC with plain FT and untagged inference. Table~\ref{tab:seed-replication} uses the same BC reference for each VLA and training seed: we compute $100(\mathrm{SR}_A-\mathrm{SR}_B)/\mathrm{SR}_{\mathrm{BC}}$ within each seed, then report the mean and sample SD. Thus, row differences are expressed relative to BC, including comparisons between two non-BC configurations. The relative gains in Fig.~\ref{fig:overview} use the same calculation for episode conditioning versus BC. Training-seed SDs describe variation across three runs, not confidence intervals or significance tests.

\subsection{Real-world experiment details}
\label{app:real-training}
\label{app:real-evaluation}
The Franka FR3 setup uses a Robotiq gripper, ZED~2i external camera, and ZED~Mini wrist camera. Pretraining and FT use 20K and 16K updates, respectively, with batch 64 and AdamW cosine learning rates from $2.5\times10^{-5}$ to $2.5\times10^{-6}$. Pretraining uses LoRA for language/action modules and a frozen vision encoder; FT makes the vision encoder trainable. The learning-rate schedules use 1,000 and 1,600 warmup updates for pretraining and FT, respectively. Both stages use fixed DROID normalization statistics without re-estimation, $224\times224$ center crops, and a 15-action prediction horizon. Simulation observations are rendered at $640\times360$ with workspace-aligned backgrounds and camera views. Every checkpoint includes the same normalization statistics for training and inference. Pretraining applies color jitter, blur, posterization, sharpness, and grayscale augmentation following MolmoB0T~\cite{deshpande2026molmob0t}.

Conditioning assigns episode-score quintiles to 17,758 demonstrations and neutral tag~3 to 1,492 unscorable episodes, without pretraining tag dropout. The ten real-world tasks cover nesting, stacking, alignment, rotation, pressing, closing, insertion, and extraction. Evaluation uses 12 layouts per task and two trials per layout; equal trial counts make task-averaged and pooled success rates identical.

\bibliographystyle{axisIEEEtran}
\bibliography{reference}
\end{document}